\documentclass[10pt,twocolumn,letterpaper]{article}

\usepackage{cvpr}
\usepackage{times}
\usepackage{epsfig}
\usepackage{graphicx}
\usepackage{amsmath}
\usepackage{amssymb}

\usepackage{algorithm}
\usepackage{algorithmic}
\usepackage{bm}
\usepackage[subrefformat=parens]{subcaption}
\usepackage{comment} 

\usepackage{color} 
\usepackage{xcolor} 

\newcommand{\yoshiapr}[1]{\textcolor{black}{#1}}
\newcommand{\yoshinew}[1]{\textcolor{black}{#1}}

\newcommand{\yoshi}[1]{\textcolor{black}{#1}}
\newcommand{\kaw}[1]{\textcolor{black}{#1}}
\newcommand{\rk}[1]{\textcolor{black}{#1}} 
\newcommand{\yoshiaapr}[1]{\textcolor{black}{#1}}
\newcommand{\ari}[1]{\textcolor{black}{#1}}
\newcommand{\arm}[1]{\textcolor{black}{#1}}
\newcommand{\yoshifinal}[1]{\textcolor{black}{#1}}

\usepackage[pagebackref=true,breaklinks=true,letterpaper=true,colorlinks,bookmarks=false]{hyperref}

\cvprfinalcopy 

\def\cvprPaperID{59} 

\ifcvprfinal\fi
\begin{document}

\title{Differentiating Objects by Motion: 
\\ Joint Detection and Tracking of Small Flying Objects}

\author{\hspace{8mm}Ryota Yoshihashi\\
\and
\hspace{-6mm}Tu Tuan Trinh\\
\hspace{-4mm}The University of Tokyo\\
\and
\hspace{-2mm}Rei Kawakami\\
\and
\hspace{0mm}Shaodi You\\
\hspace{0mm}CSIRO-Data61\\
\hspace{0mm}Australian National University\\
\and
\hspace{-38mm}Makoto Iida\\
{\hspace{2mm} The University of Tokyo}\\
\and
\hspace{-24mm}Takeshi Naemura\\
\and
{\vspace{-1mm} \tt\small \{yoshi, tu, rei, naemura\}@hc.ic.i.u-tokyo.ac.jp }\\
{\vspace{-1mm} \tt\small iida@ilab.eco.rcast.u-tokyo.ac.jp}\\  
{\vspace{-1mm} \tt\small shaodi.you@data61.csiro.au}\\
}

\makeatletter
\let\@oldmaketitle\@maketitle
\renewcommand{\@maketitle}{\@oldmaketitle
  \vspace{-8mm} \hspace{3mm} \includegraphics[width=480pt]{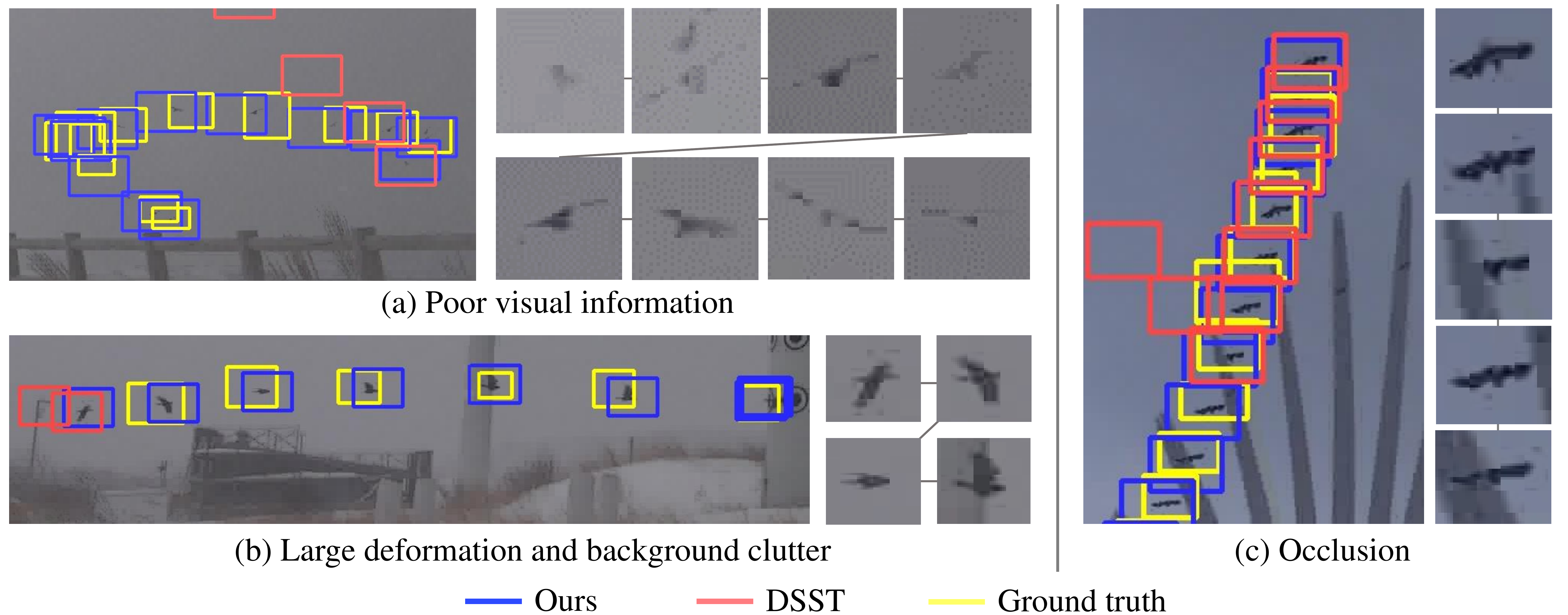}
  {\\ Figure 1: Importance of multi-frame information for recognizing apparently small \ari{flying} objects \yoshiaapr{(birds in these examples)}. While visual features in single frames are vague and limited, multi-frame information, including \rk{deformation and pose changes}, provides better clues with which to recognize birds. 
 \yoshifinal{To extract such useful motion patterns, tracking is necessary for compensating translation of objects, but the tracking itself is a challenge due to the limited visual information.}
  The blue boxes are birds tracked by our method that
  utilizes multi-frame representation for detection, while the
  red boxes are the results of a \ari{single-frame} handcrafted-feature-based tracker~\cite{danelljan2017discriminative} 
  \ari{, which tends to fail when tracking small objects.}
  }
  \vspace{-0mm}
    \bigskip}
\makeatother
\setcounter{figure}{1}

\maketitle

\begin{abstract} \vspace{-4mm}
\yoshinew {While generic object detection has achieved large improvements with rich feature hierarchies from deep nets, detecting small objects with poor visual cues remains challenging.
Motion cues from multiple frames may be more informative for detecting such hard-to-distinguish objects in each frame. However, how to encode discriminative motion patterns, such as \rk{deformations and pose changes} that characterize objects, has remained an open question.
To learn them and thereby realize small object detection,} we present a neural model called the {\it Recurrent Correlational Network}, where detection and tracking are \rk{jointly} 
performed over a multi-frame representation learned through a single, trainable, and end-to-end network. A convolutional long short-term memory network is utilized for learning informative \rk{appearance change} for detection, while learned representation is shared in tracking for enhancing its performance. In experiments with datasets \arm{containing images of scenes with} small flying objects, such as birds and unmanned aerial vehicles, the proposed method yielded consistent improvements in detection performance over deep single-frame detectors and existing motion-based detectors. Furthermore, our network performs \arm{as well as} state-of-the-art generic object trackers when it was evaluated as a tracker on the bird dataset.
\end{abstract} \vspace{-5mm}

\section{Introduction}
\yoshiapr{
\kaw{Detection of visually small objects is often required in wide-area surveillance~\cite{collins2000system,coluccia2017drone,rozantsev2017detecting}.
Rich visual \arm{representations} by deep convolutional networks (convnets)~\cite{girshick2014rich} pre-trained on a large-scale still-image dataset~\cite{ILSVRC15} \arm{are of limited use on such} objects, because they \arm{appear} blurred and textureless \arm{in images} owing to their \arm{small} apparent size.
For them to be detected, motion, namely the} changes in their temporal appearance over a longer time frame, may offer richer information than appearance at a glance. As shown in Fig.~1, a bird is much easier to identify when multiple frames are available. 
}
\kaw{However, \arm{it remains unclear} how to learn motion features that are powerful enough to differentiate object. }

\kaw{In this paper, we present a method that exploits motion cues for small object detection. 
\arm{Although}} we utilize learnable pipelines based on convolutional and recurrent networks, our key idea is \kaw{letting the network \yoshiaapr{focus on} informative deformations such as flapping of wings to differentiate target objects for detection, while removing less useful translations~\cite{park2013exploring} by simultaneously tracking them with the learned visual representation.}
\arm{To make this possible,} 
our framework performs joint detection and tracking. 
It utilizes convolutional long short-term memory (ConvLSTM)~\cite{xingjian2015convolutional} 
to learn a discriminative multi-frame representation for detection,
while it also enables correlation-based tracking over its output. Tracking is aided by the shared representation afforded by the training of the detector, and the overall framework is simplified \arm{because there are fewer} parameters to be learned.
\yoshiapr{We refer to the pipeline as {\it Recurrent Correlational Network}.}
\kaw{Experiments on single-class, fully supervised small object detection in videos 
targeting birds~\cite{trinh2016} and unmanned aerial vehicles (UAVs)~\cite{rozantsev2017detecting} show consistent improvements by our network over single-frame baselines and previous multi-frame methods. When evaluated as a tracker, ours also outperforms 
existing hand-crafted-feature-based and deep generic-object trackers in the bird dataset.}

\yoshiapr{
Our contribution is three-fold. First, we show that 
motion patterns learned via ConvLSTM improves detection performance in small object detection.
\arm{Our network} outperforms single-frame baselines, score-averaging baselines, and existing multi-frame methods in flying-object datasets, which indicates the importance of motion cues in these domains.
Second, we introduce a novel framework for simultaneous object detection and tracking in video, which efficiently handles motion learning. 
This is the first {\yoshinew recurrent model} to achieve joint detection and tracking with deep learning. 
Third, our network is accurate when evaluated as a separate tracker in the dataset where class-specific detectors can be trained. The proposed network outperforms existing trackers based on various hand-crafted features, and performs slightly better or on par \arm{with} convnet-based trackers. 
\yoshifinal{Our results gives a prospect toward domain-specific multi-task representation learning, which should open up application fields that 
generic detectors or trackers do not directly generalize.}
The relevant code and data will be published upon acceptance of this paper.
}

\begin{figure*}[t]
  \begin{center}
    \includegraphics[width=475pt]{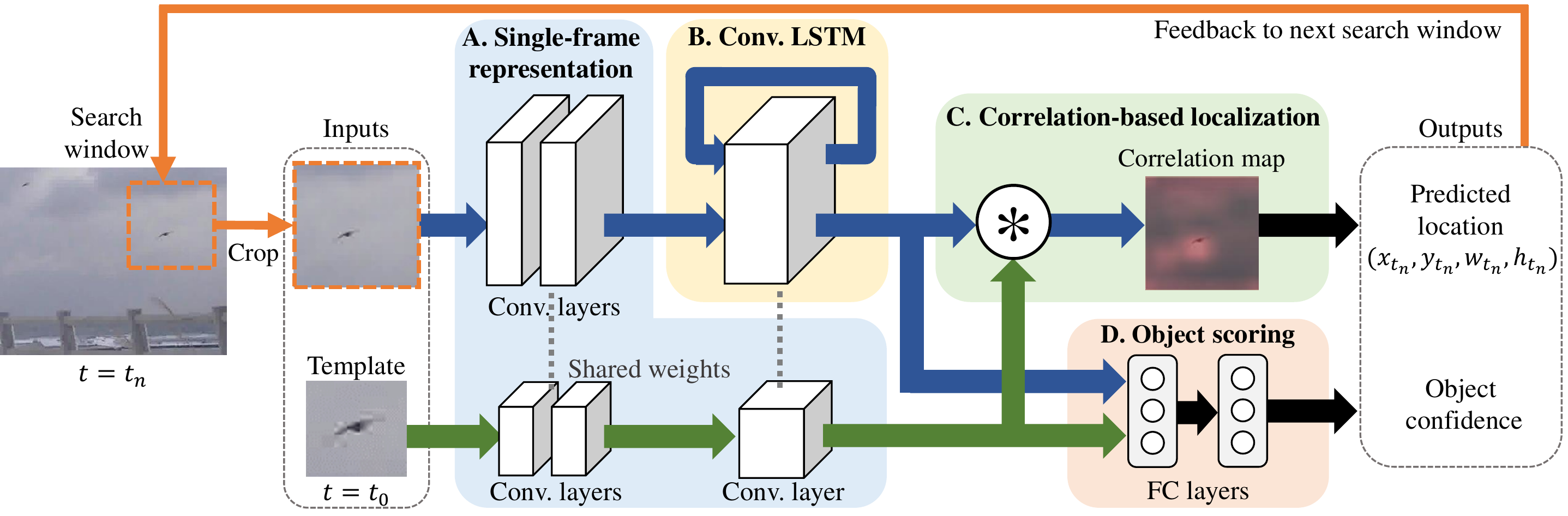}
    \vspace{-7mm}
  \end{center}
  \caption{Overview of the proposed network, called {\it Recurrent Correlation Network} (RCN). It consists of \yoshifinal{the four modules}: \yoshiaapr{Convolutional layers for single-frame representations (A), ConvLSTM layers for multi-frame representations (B), cross-correlation layers for localization (C), and fully-connected layers for object scoring (D)}. Green arrows show the information stream from templates \yoshiapr{(the proposals in the first frame at $t=t_0$)}, and blue arrows show that from search windows.}
  \vspace{-5mm}
  \label{fig:net}
\end{figure*}

\vspace{-2mm}
\section{Related work}
\vspace{-1.5mm}
\noindent {\bf Small object detection} \hspace{1.5mm}
\yoshiapr{Detection of small objects has been
tackled in \arm{the} surveillance community~\cite{collins2000system}, and
recently has attracted much attention \arm{since the advent} of UAVs~\cite{coluccia2017drone,schumann2017deep}.
\rk{\ari{Small} pedestrians~\cite{bunel2016detection} and faces~\cite{hu2016finding} have also been considered, and some recent studies try to detect small common objects in generic-object detection setting~\cite{chenr, li2017perceptual}. Studies are more focused on scale-tuned convnets with moderate depths and a wider field of view, and despite of its importance, motion has not yet been incorporated in \ari{these} domains.}
}
\vspace{1mm}

\noindent {\bf Object detection in video} \hspace{1.5mm}
Having achieved significant success in generic object detection in still images~\cite{girshick2014rich,girshick2015fast,ren2015faster,liu2016ssd,dai2016r,redmon2016yolo9000}, the \ari{research trends have} begun moving toward efficient generic object detection in videos~\cite{ILSVRC15}. \yoshiapr{The video detection task poses new challenges, such as how to process voluminous video data efficiently and 
how to handle appearance of objects differing from still images
due to rare poses ~\cite{feichtenhofer2017detect, zhu2017flow}.}
Very recent studies have begun improving on detection in videos; examples include T-CNNs~\cite{kang2017t,kang2016tubelets} that use trackers for propagating high-confidence detection, 
and deep feature flow~\cite{zhu2016deep} and flow-guided feature aggregation~\cite{zhu2017flow} that involves feature-level smoothing using optical flow. 
\yoshinew {One of the closest idea to ours is joint detection
and bounding-box linking by coordinate regression~\cite{feichtenhofer2017detect}.
\yoshiapr{These models \arm{that have been used} in ILSVRC-VID are more like modeling temporal consistency than understanding motion. Thus, it \arm{remains unclear} whether or how inter-frame information extracted from motion or deformation aids in understanding objects}.
}
{\yoshinew In addition, they all are based on popular convolutional generic still-image detectors~\cite{dai2016r,girshick2015fast,girshick2014rich,liu2016ssd,redmon2016yolo9000,ren2015faster} and it is not clear to what extent such generic object detectors, which are designed for and trained in dataset collected from the web, generalize to task-specific datasets~\cite{dollar2012pedestrian,hosang2015taking,zhang2016faster}. In the datasets for flying objects detection that we use~\cite{trinh2016,rozantsev2017detecting}, the domain gap is especially large due to differences in the appearance of objects and backgrounds, \ari{as well as} scale of objects. 
Thus, we decided to use simpler region proposals and fine-tune our network as region classifiers in each dataset.
}

\vspace{1mm}
\noindent {\bf Deep trackers} \hspace{1.5mm}
Recent studies have intensively examined convnets and recurrent nets for tracking. Convnet-based trackers learn convolutional layers to acquire rich visual representation. 
Their localization strategies are diverse, \yoshiapr{including  classification-based~\cite{nam2016learning}, similarity-learning-based~\cite{limulti}, regression-based~\cite{held2016learning}, and  correlation-based~\cite{bertinetto2016fully,valmadre2017end} approaches. 
While classification of densely sampled patches~\cite{nam2016learning}
is the most accurate \arm{in} generic benchmarks, its computation is slow and regression-based one~\cite{held2016learning} and correlation-based ones~\cite{bertinetto2016fully,valmadre2017end} are used \arm{instead in} real-time.
\rk{Our network also incorporates a correlation-based localization mechanism, having its performance enhanced by the representation shared by the detector.}
}

Recurrent nets~\cite{werbos1988generalization,hochreiter1997long} efficiently handle temporal structures in sequences \yoshiapr{and thus, they have been used for} tracking~\cite{ning2016spatially,gordon2017re3,milan2017online,wang2017trajectory}.  However, most utilize separate convolutional and recurrent layers, and have a fully connected recurrent layer, which may lead to a loss of spatial information. 
\yoshiaapr{Thus, currently recurrent trackers do not perform as well as the best single-frame convolutional trackers in generic benchmarks.}
\ari{One study used ConvLSTM with simulated robotic sensors for \rk{handling occlusion}} ~\cite{ondruska2016deep}.

\vspace{1mm}
\noindent {\bf Joint detection and tracking} \hspace{1mm} 
The relationship between object detection and tracking is a long-term problem in itself; before the advent of deep learning, it had only been explored with classical tools.
\rk{In the track--learn--detection (TLD) framework~\cite{kalal2012tracking}, a trained} detector enables long-term tracking by re-initializing trackers after temporal disappearance of objects. 
Andriluka \etal uses a single-frame part-based detector and shallow unsupervised learning based on temporal consistency~\cite{andriluka2008people}. 
Tracking by associating detected bounding boxes \cite{huang2008robust} is another popular approach. However, in this framework, recovering undetected objects is challenging because tracking is more akin to post-processing following detection than to joint detection and tracking.

\vspace{1mm}
\noindent {\bf Motion feature learning} \hspace{1.5mm}
\rk{Motion feature learning, and hence the use of recurrent nets, are more active in video classification~\cite{karpathy2014large} and action recognition~\cite{soomro2012ucf101}.}
\yoshiapr{Studies have shown that \arm{LSTMs yield improvement in accuracy}~\cite{wang2015action,weinzaepfel2015learning,donahue2015long}.
For example, VideoLSTM~\cite{li2016videolstm} uses the idea of inter-frame correlation to recognize actions with attention. \rk{However, with \ari{action recognition} datasets, the networks may not fully utilize human motion features apart from appearance, backgrounds and contexts ~\cite{he2016human}.}}

\yoshiaapr{
Optical flow~\cite{lucas1981iterative,horn1981determining,dosovitskiy2015flownet} is a pixel-level alternative to trackers
to describe motion~\cite{park2013exploring,gladh2016deep,zhu2016deep,zhu2017flow}.
}
{\yoshinew Accurate flow estimation is, however, challenging in small flying
object detection tasks due to the small apparent size of the targets
and the large inter-frame disparity by fast motion~\cite{rozantsev2017detecting}.}
While we focus on high-level motion stabilization and motion-pattern learning via tracking, we believe flow-based low-level motion handling is orthogonal and complementary to ours depending on the application areas.

\vspace{-2mm}
\section{Recurrent Correlational Networks}
\vspace{-1mm}
\yoshiapr{To exploit motion information via simultaneous detection and tracking, we present \arm{the} {\it Recurrent Correlational Networks} as shown in Fig.~\ref{fig:net}}. 
\arm{The network} consists of four modules: (A) convolutional layers, (B) ConvLSTM layers, (C) a cross-correlation layer, and (D) fully connected layers for object scoring. First, the convolutional layers model single-frame appearances of target and non-target regions, including other objects and backgrounds. 
\yoshiaapr{
Second, the ConvLSTM layers encode temporal sequences of
single-frame appearances, and extract the discriminative motion patterns.}
Third, the cross-correlation layer convolves the \rk{representation} of the template to that of search windows in subsequent frames, and generates correlation maps that are useful for localizing the targets. Finally, the confidence scores of the objects are calculated with fully-connected layers based on the multi-frame representation.
\yoshifinal{
The network is supervised by the detection loss,
and the tracking gives locational feedback for region of interest in next frames during training and testing.
}

{\yoshinew
Our detection pipeline is based on region
proposal and classification of the proposal,
as in region-based CNNs~\cite{girshick2014rich}.
The main difference is in that our joint detection and tracking network simultaneously track the given proposals in the following frames, and the results
of the tracking are reflected in the classification scores, that are used as detectors' confidence scores. 
}

\vspace{1.5mm}
\noindent {\bf Convolutional LSTM} \hspace{1.5mm}
In our framework, the ConvLSTM module~\cite{xingjian2015convolutional} is used for motion feature extraction (Fig.~\ref{fig:net}~B). It is a convolutional counterpart of LSTM~\cite{hochreiter1997long}. It replaces inner products in the LSTM with convolution, \yoshifinal{and this is more suitable for motion learning, since the network is more sensitive to local spatio-temporal patterns rather than the global patterns.}
It works as a sequence-to-sequence predictor; specifically, it
takes series $(x_1, x_2, x_3, ..., x_t)$ of single-frame representations whose length is $t$ as input, and outputs a merged single representation $h_t$, 
at each timestep $t = 1, 2, 3, ..., L$.

For the sake of completeness, we show the formulation of ConvLSTM below. 
\begin{eqnarray} \label{eqn:lstm} 
i_t &=& \sigma(w_{xi} * x_t + w_{hi} * h_{t - 1} +b_i) \nonumber \\ 
f_t &=& \sigma(w_{xf} * x_t + w_{hf} * h_{t - 1} + b_f) \nonumber \\
c_t &=& f_t \circ c_{t - 1} + i_t \circ \mathrm{tanh}(w_{xc}  * x_t + w_{hc}  \circ h_{t - 1} + b_c) \nonumber \\
o_t &=& \sigma(w_{xo} * x_t + w_{ho} * h_{t - 1} +　b_o) \nonumber \\
h_t &=& o_t \circ \mathrm{tanh}(c_t).
\vspace{4cm}
\end{eqnarray}
Here, $x_t$ and $h_t$ denote the input and output of the layer at timestep $t$, respectively. The states of the memory cells are denoted by $c_t$. $i_t$, $f_t$, and $o_t$ and are called gates, which work for selective memorization. `$\circ$' denotes the Hadamard product.
ConvLSTM is also well suited to exploit the spatial correlation for joint tracking, since its output representations are in 2D.

While ConvLSTM is effective at video processing, it inherits the complexity of LSTM. The gated recurrent unit (GRU) is a simpler alternative to LSTM that has fewer gates, and it is empirically easier to train on some datasets~\cite{chung2015gated}. A convolutional version of the GRU (ConvGRU)~\cite{siam2016convolutional} is as follows: 
\begin{eqnarray}
z_t &=& \sigma(w_{xz} * x_t + w_{hz} * h_{t - 1} + b_z) \nonumber \\ 
r_t &=& \sigma(w_{xr} * x_t + w_{hr} * h_{t - 1} + b_r)  \\ 
h_t &=& z_t \circ h_{t-1} + (1 - z_t) \nonumber \\ 
	& & \circ \hspace{1mm} \mathrm{tanh}(w_{xh} * x_t + w_{hh} * (r_t \circ h_{t - 1}) + b_h) .\nonumber 
\vspace{4cm}
\end{eqnarray}
ConvGRU has only two gates, \yoshiapr{namely an update gate $z_t$ and reset gate $r_t$}, while ConvLSTM has three.
ConvGRU can also be incorporated into our pipeline; later we provide an empirical comparison between ConvLSTM and ConvGRU.

\begin{figure}[t]%
  \begin{center}
    \includegraphics[width=235pt]{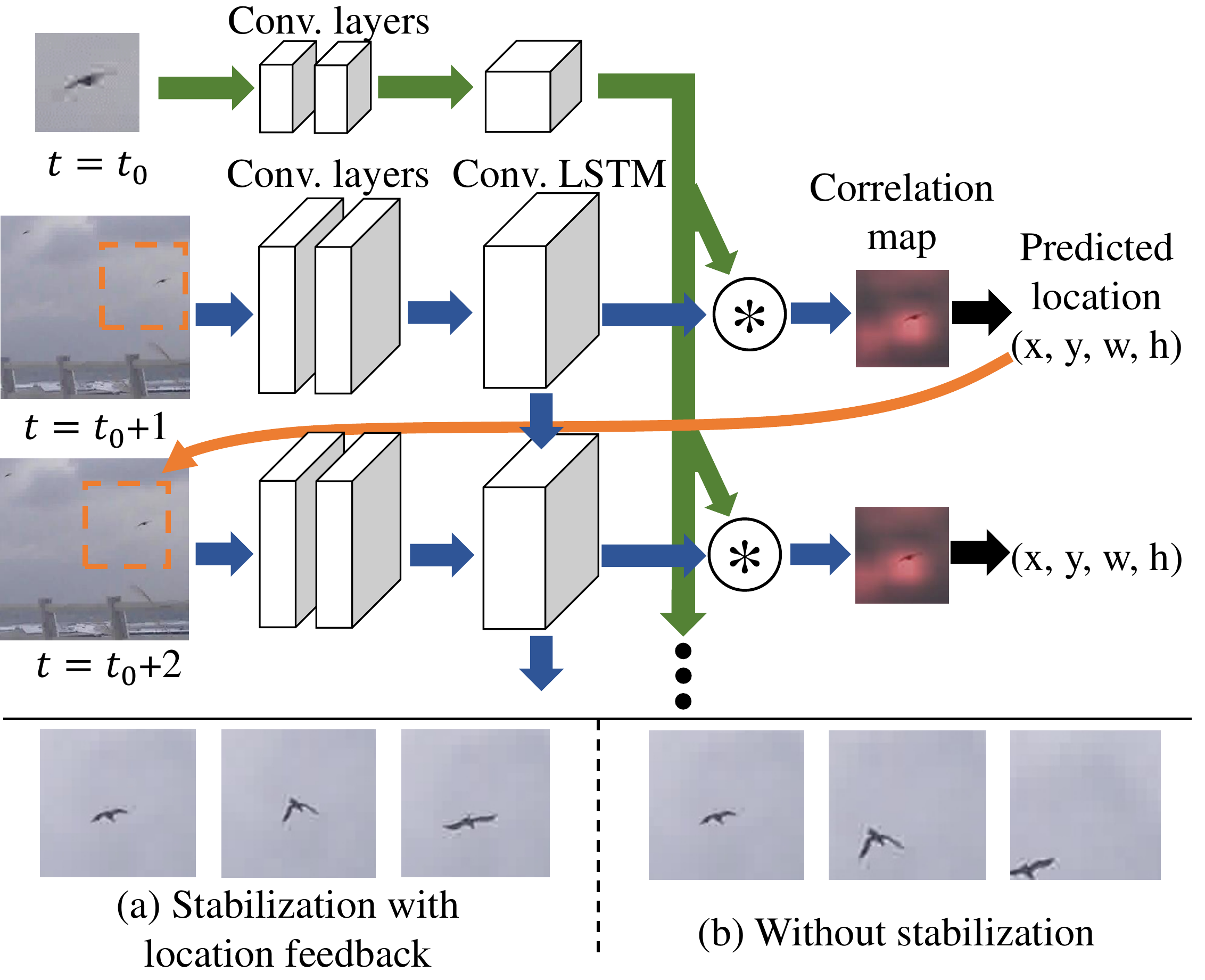}
    \vspace{-10mm}
  \end{center}
  \caption{Temporal expansion of the proposed network. The joint tracking is incorporated as part of the feedback in the recurrent cycle. 
  \yoshifinal{This feedback provides stabilized observation of moving objects (a), 
  while learning from deformation is difficult without stabilization (b).}} 
  \label{fig:tex}
  \vspace{-4mm}
\end{figure}

\vspace{1.5mm}
\noindent {\bf Correlation-based localization} \hspace{1.5mm}
\yoshifinal{
The correlation part (Fig.~\ref{fig:net}~C) aims to stabilize moving objects' appearance by tracking.
The localization results are fed back to the next
input, as shown in Fig.~\ref{fig:tex}. 
This feedback allows ConvLSTM to learn deformations and pose changes apart from translation 
(Fig.~\ref{fig:tex} a), while local motion patterns
are invisible due to translation without stabilization (Fig.~\ref{fig:tex} b).
}

Cross-correlation is an operation that relates two inputs, and outputs a correlation map that indicates how similar each patch in an image is to another. It is expressed as
\begin{equation}
\vspace{-1mm}
C(\bm{p}) = \bm{f} * \bm{h} =  \sum_{\bm{q}} \bm{f}(\bm{p} + \bm{q})  \cdot \bm{h}(\bm{q}). \\ \vspace{-1mm}
\end{equation}　
where $\bm{f}$ and $\bm{h}$ denote the multi-dimensional feature representations of the search window and template, respectively. $\bm{p}$ is for every pixel's coordinates in the domain of $\bm{f}$, and $\bm{q}$ is for the same but in the domain of $\bm{h}$. Two-dimensional (2D) correlation between a target patch and a search window is equivalent to densely comparing the target patch with all possible patches within the search window.  The inner product is used here as a similarity measure. 

In the context of convolutional neural networks, the cross-correlation layers can be considered to be differentiable layers without learnable parameters; namely, a cross-correlation layer is a variant of the usual convolutional one whose kernels are substituted by the output of another layer. Cross-correlation layers are bilinear with respect to two inputs, and thus are differentiable. The computed correlation maps are used to localize the target by 
\vspace{-0mm}
\begin{equation}
\bm{p}_{target} = \mathrm{argmax}_{\bm{p}} C(\bm{p}) 
\vspace{-0mm}
\end{equation} 

\vspace{-2mm}
\paragraph{Single-frame representation}
A multi-layer convolutional representation is inevitable in natural image recognition, although the original ConvLSTM ~\cite{xingjian2015convolutional} did not use non-recurrent convolutional layers in radar-based tasks. Following recent tandem CNN-LSTM models for video recognition~\cite{donahue2015long}, we insert non-recurrent convolutional layers before the ConvLSTM layers (Fig.~\ref{fig:net}~A). 
Arbitrary covolutional architectures can be
incorporated and we should choose the proper
ones for each dataset.
We experimentally tested two different structures of varying depth.  

We need to extract an equivalent representation from the object template for the search windows. For this, we use ConvLSTM, in which the recurrent connection is severed. Specifically, we force the forget gates to be zero and  enter zero vectors instead of the previous hidden states. This layer is equivalent to a convolutional layer with $tanh$ nonlinearity and sigmoid gates. It shares weights with $w_{xc}$ in Eq. \ref{eqn:lstm}.
\vspace{-2mm}
\paragraph{Search window strategy}
In object tracking, as the speed of the target objects is physically limited, limiting the area of the search windows, where the correlations are computed, is a natural way to reduce computational costs. We place windows the centers of which are at the previous locations of \arm{the} objects, \yoshiapr{with a radius $R = \alpha \max(W, H)$, where $W$ and $H$} are  the width and height of the bounding box of the candidate object. We then compute the correlation map for windows around each candidate object. 
We empirically set the size of the search windows to $\alpha = 1.0$.
The representation extracted from the search windows is also fed to the object scoring part of the network, which yields large field-of-view features and provides contextual information for detection. 

\vspace{-4mm}
\paragraph{Object scoring}
For object detection, the tracked candidates need to be scored according to likeness. We use fully connected (FC) layers for this purpose (Fig.~\ref{fig:net}~D). We feed both the representations from the templates (green lines in Fig.~\ref{fig:net}) and the search windows (blue lines in Fig.~\ref{fig:net}) into the FC layers by concatenation. We use two FC layers, where the number of dimensions in the hidden vector was 1,000. 

We feed the output of each timestep of ConvLSTM into the FC layers and average the scores. In theory, the representation of the final timestep after feeding the last frame of the sequence should provide the maximum information.  However, we found that the average scores are more robust in case of tracking failures or the disappearance of targets.

\vspace{-4mm}
\paragraph{Training}
Our network is trainable with ordinary gradient-based optimizers in an end-to-end manner because all layers are differentiable. 
\yoshifinal{We separately train convolutional parts
and ConvLSTM to ensure fast convergence and avoid overfitting.} 
We first initialize single-frame-based convnets by pre-trained weights in the ILSVRC2012-CLS dataset, the popular and largest generic image dataset. We then fine-tune single-frame convnets in the target datasets (birds and drones) without ConvLSTM. Finally, we add the convolutional LSTM, correlation layer, and FC layers to the networks and fine-tune them again.  For optimization, we use the SGD solver of Caffe~\cite{jia2014caffe}.  In the case reported here, the total number of iterations was 40,000 and the batch size was five. The original learning rate was 0.01, and was \yoshiapr{reduced by a factor of 0.1} per 10,000 iterations. The loss was the usual sigmoid cross-entropy \yoshifinal{for detection.} 
We freeze the weights in the pre-trained convolutional layers after connecting  to the convolutional LSTM to avoid overfitting. 

\yoshifinal{During training of ConvLSTM, we use pre-computed trajectories predicted by a
single-frame convolutional tracker, which consists of the final convolutional layers of the pre-trained single-frame convnet and a correlation layer. They are slightly inaccurate but have similar trajectories to those of our final network. 
Then, }we store cropped search windows in the disk during training for efficiency, to reduce disk access by avoiding the re-cropping of the regions of interest out of the 4K-resolution frames during training.
During the testing phase, the network observes trajectories \rk{estimated} by itself, which are different from the ground truths that are used in the training phase.  
This training scheme is often referred to as teacher forcing~\cite{williams1989learning}.  Negative samples also need trajectories in training, but we do not have their ground truth trajectories because only the positives are annotated in the detection datasets.  

\vspace{-2mm}
\section{Experiments}

\hspace{-2mm}
\begin{figure*}[t]
	\hspace{-3mm}
	\begin{tabular}{c|c}
    \begin{minipage}{0.66\hsize}\vspace{-3mm}
    	\begin{minipage}{0.5\hsize}
        	\hspace{-5mm}
        	\includegraphics[width=180pt]{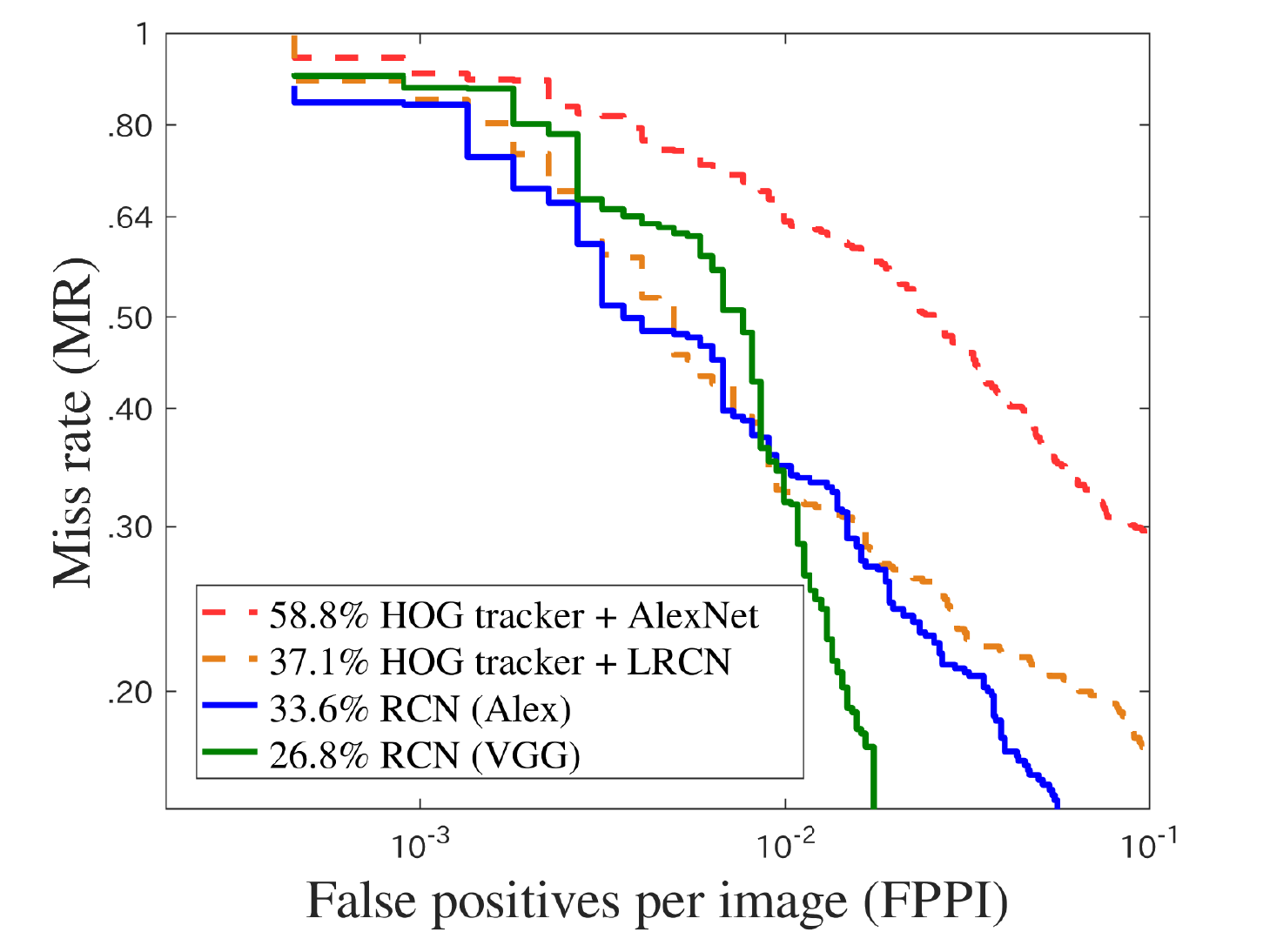}
            \captionsetup{labelformat=empty,labelsep=none}
            \vspace{-8mm}
            \caption{\small {\it Reasonable} subset (40 pixels --)}
        \end{minipage}
        \begin{minipage}{0.5\hsize}
        	\hspace{-5mm}
        	\includegraphics[width=180pt]{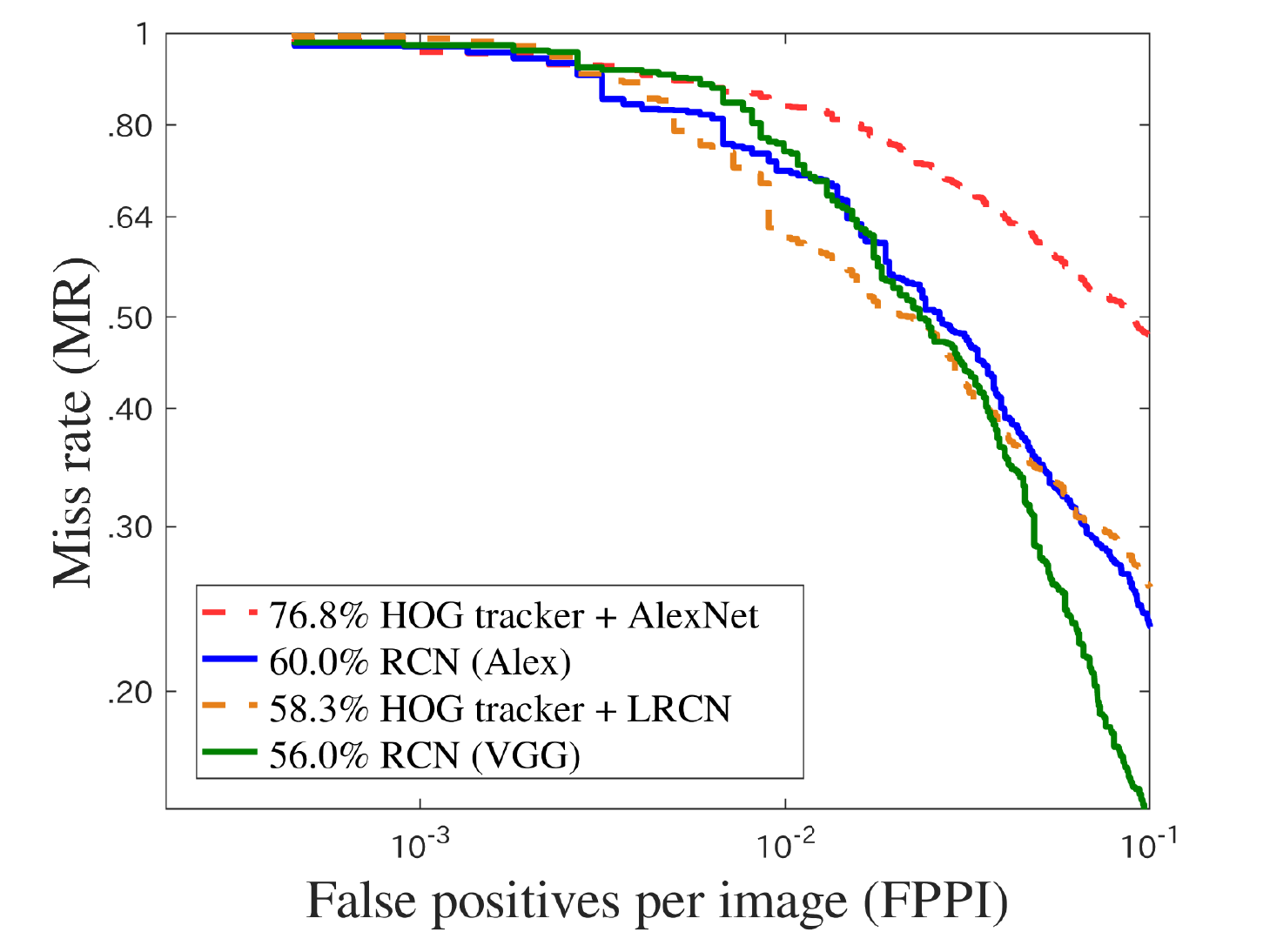}
            \captionsetup{labelformat=empty,labelsep=none}
            \vspace{-8mm}
            \caption{\small {\it Small} subset (--40 pixels)}
        \end{minipage}
        \\
        \begin{minipage}{0.5\hsize}
        	\hspace{-5mm}
            \vspace{0mm}
        	\includegraphics[width=180pt]{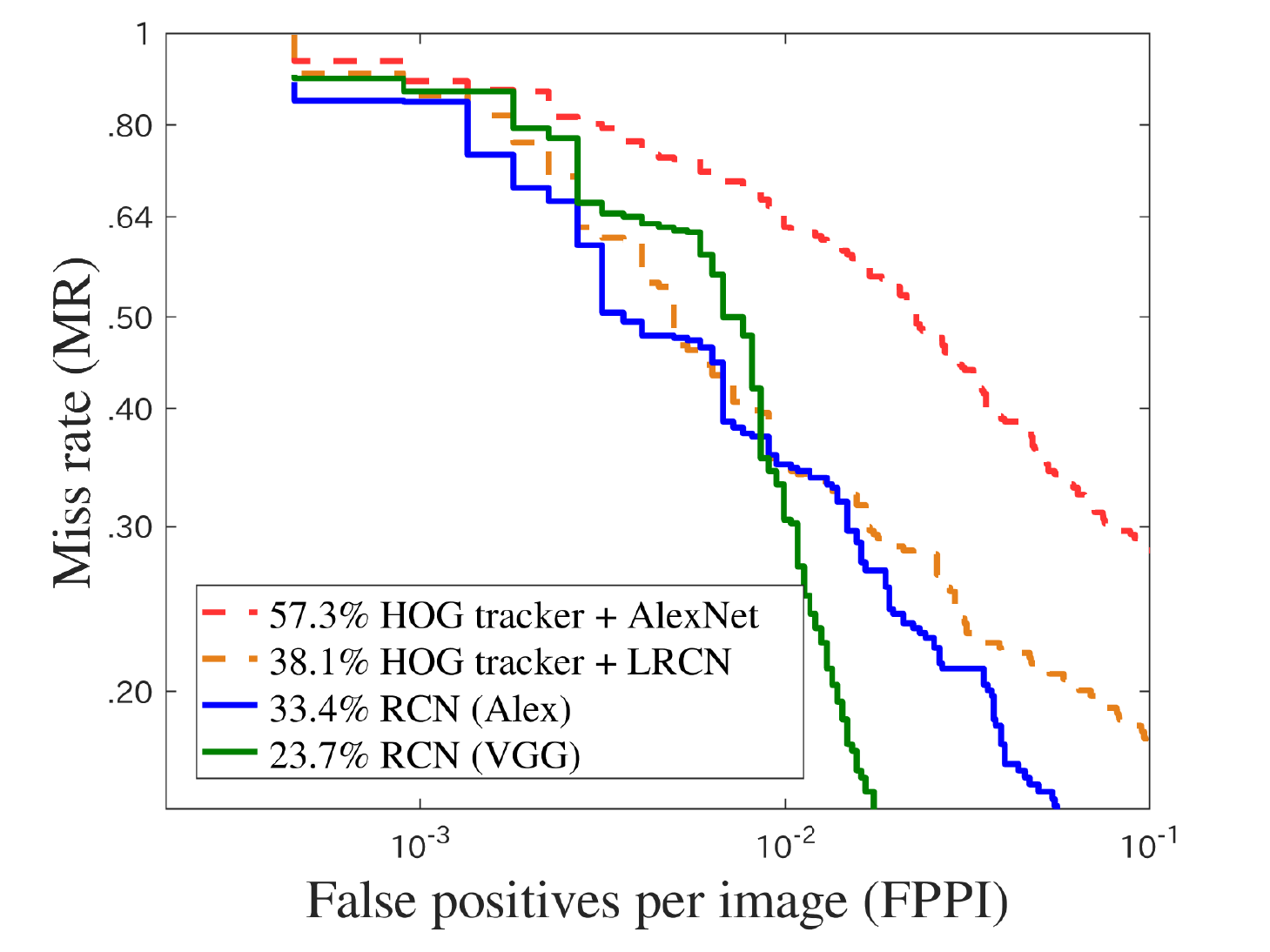}
            \captionsetup{labelformat=empty,labelsep=none}
            \vspace{-8mm}
            \caption{\small {\it Middle-size} subset (40 -- 60 pixels)}
        \end{minipage}
         \begin{minipage}{0.5\hsize}
         	\hspace{-5mm}
        	\includegraphics[width=180pt]{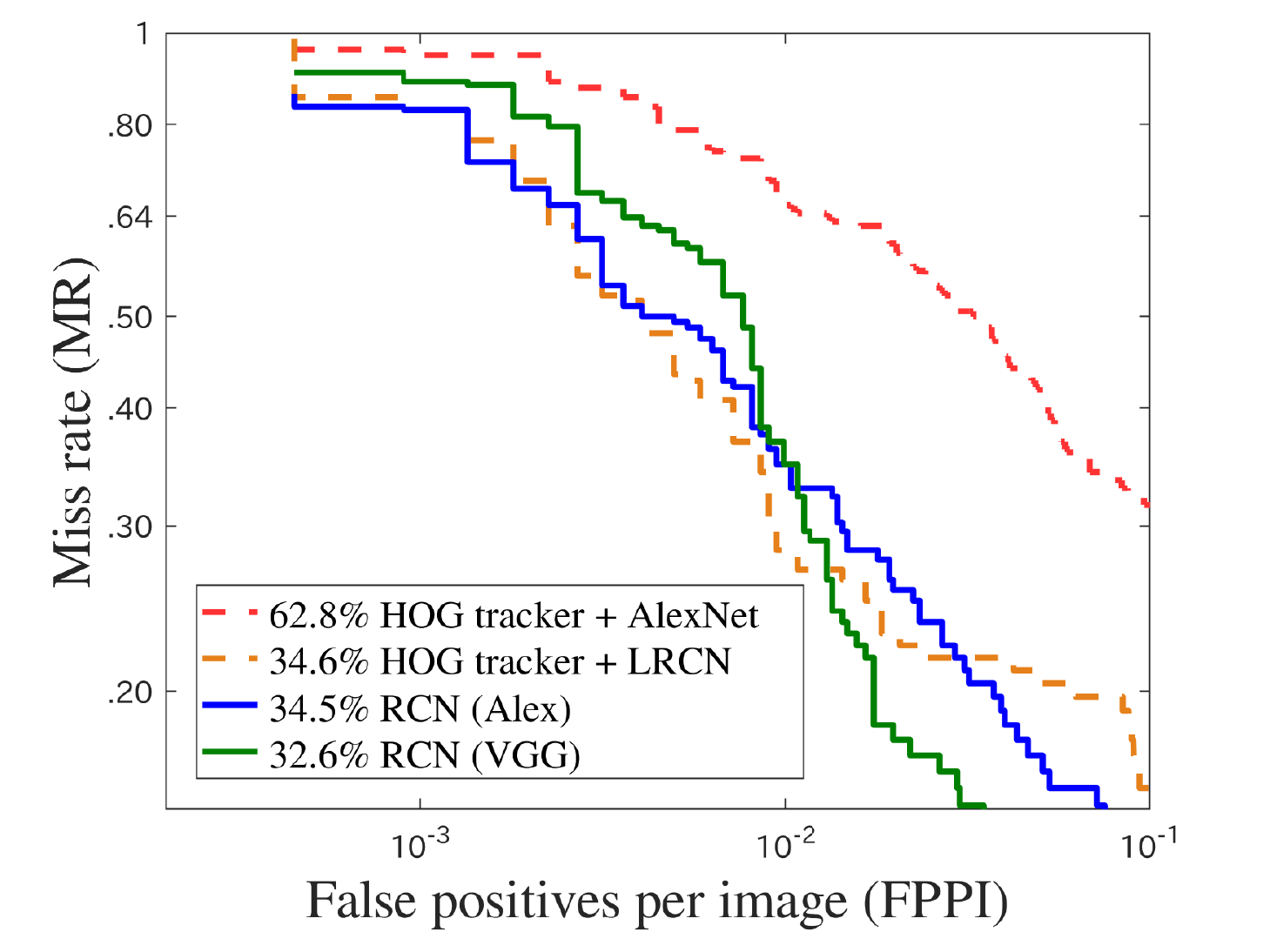}
            \captionsetup{labelformat=empty,labelsep=none}
            \vspace{-8mm}
            \caption{\small {\it Large} subset (60 pixels --)}
            \end{minipage}
        \setcounter{figure}{3}
        \caption{Detection results. The lower left is better. Our RCN (VGG) outperformed all of the other methods with deeper convolutional layers, and our RCN (Alex) outperformed the previous method with the same convolutional layer depth on three subsets. The subsets are distinguish by the sizes of birds in the images. } \vspace{-2mm}
        \label{fig:rocdet}
      \end{minipage}
      &
      \hspace{-1mm}
      \begin{minipage}{0.33\hsize}\vspace{-5mm}
      	\begin{center}
      	\begin{minipage}{1\hsize}
        	\vspace{-0mm} \hspace{-2mm}
        	\includegraphics[width=185pt]{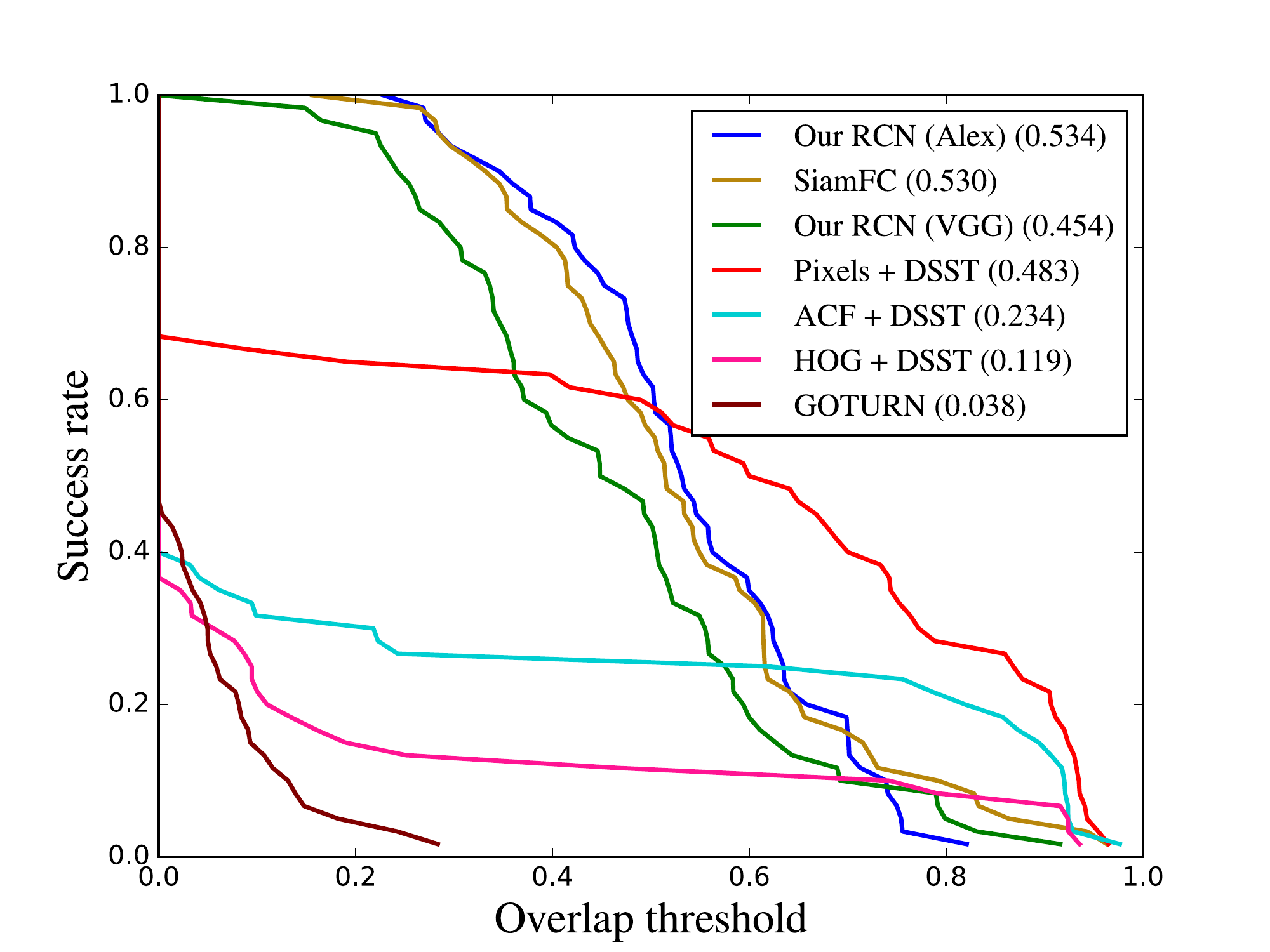}
            \vspace{-8mm} 
            \captionsetup{labelformat=empty,labelsep=none}
            \caption{\hspace{8mm} \small 30-frame snippets}
        \end{minipage}
        \\
        \begin{minipage}{1\hsize}
        	\vspace{-1mm} \hspace{-2mm}
        	\includegraphics[width=185pt]{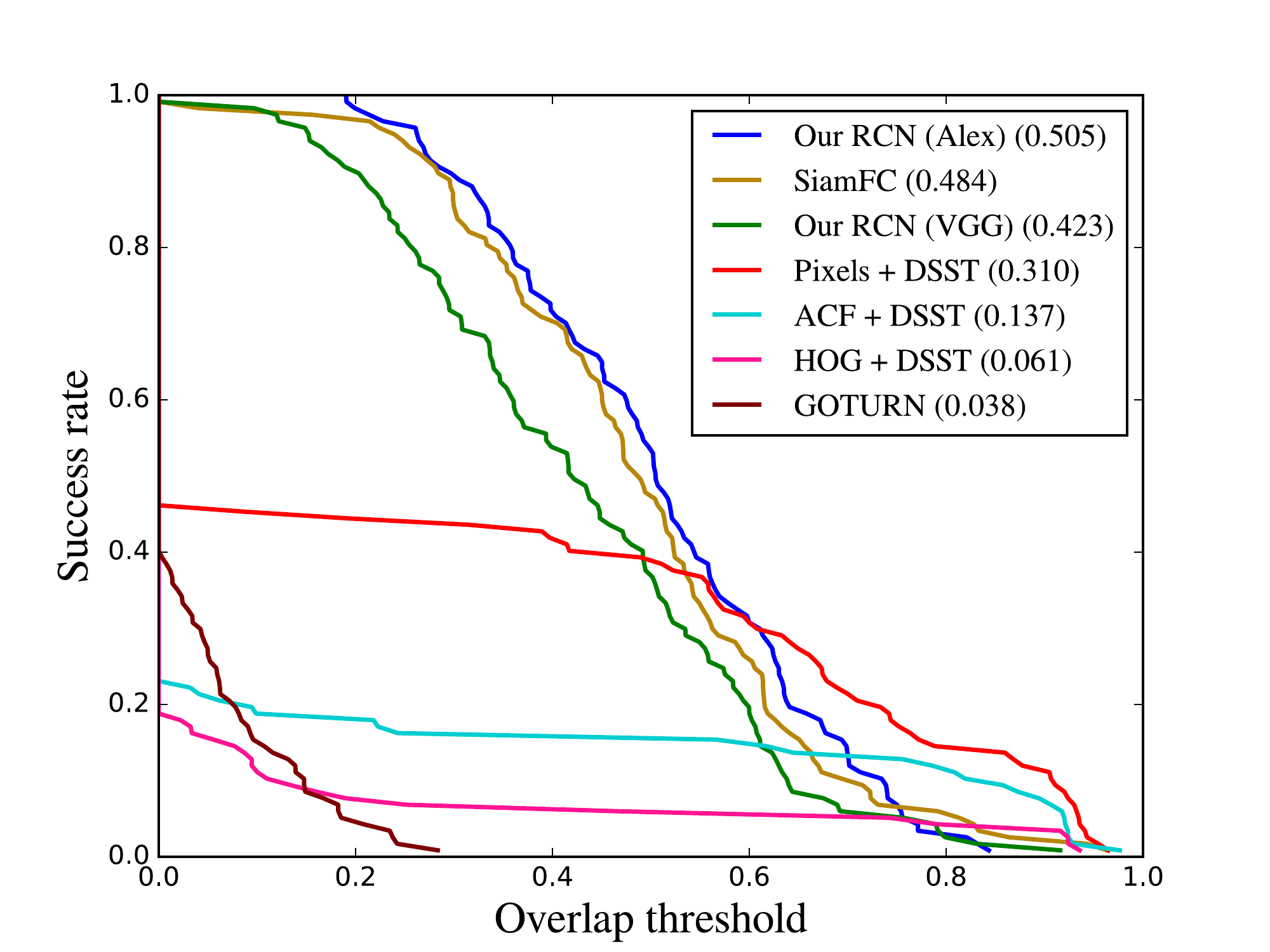}
            \vspace{-8mm} 
            \captionsetup{labelformat=empty,labelsep=none}
        \caption{\hspace{8mm} \small 60-frame snippets}
        \end{minipage}
        \setcounter{figure}{4}
        \vspace{-2mm}
        \caption{Tracking results. The upper right is better. The proposed methods outperformed DSST trackers with various handcrafted features and the ImageNet-pretrained deep trackers.} \vspace{-2mm}
        \label{fig:roctrk}
        \end{center}
      \end{minipage}
      
    \end{tabular}
    \vspace{-2mm}
    \vspace{-2mm}
\end{figure*}
The main purpose of the experiments was to investigate the performance gain owing \yoshiapr{to the learned motion patterns with joint tracking in small object detection tasks}. We also investigated the tracking performance of our method and compared it with that of \yoshiaapr{trackers with a variety of features
\ari{as well as} convolutional trackers}. 

\begin{table}[tb]
	\begin{center}
    \caption{Statistics of the datasets.}
	\vspace{-4mm}
    \small
  \begin{tabular}{|c|c|c|} \hline
  	& Bird~\cite{trinh2016} & UAV~\cite{rozantsev2017detecting} \\ \hline
   Frame resolution & 3840 $\times$ 2160  & 752 $\times$ 480 \\ 
   Ave. object resolution & 55 pixels & 18 pixels \\ 
   \#Test frames & 2,222 & 5,800 \\ 
   \#Training boxes & 10,000 & 8,000 \\ \hline
  \end{tabular}
  \label{tab:datasets}
  \end{center}
  \vspace{-9mm}
\end{table}
We first used a recently constructed video-based bird dataset~\cite{trinh2016}.
This dataset involves detecting birds around a wind farm. The resolution is 4K and the frame rate is 30 fps, which made processing the dataset a challenge due to its large volume.
The most frequent size of the birds is 55 pixel.
Although the dataset consists of images taken from a fixed-point camera, it has changes in illumination owing to the weather, changing background patterns owing to clouds, and variation in the appearance of birds due to occlusion and deformation. 
We also tested our method on a UAV dataset~\cite{rozantsev2017detecting} to see whether  it can be applied to other flying objects. This dataset consists of 20 sequences of hand-captured videos. It consists of approximately 8,000 bounding boxes of flying UAVs. All the UAVs in this dataset are multi-copters. We followed the training/testing split provided by the authors of ~\cite{rozantsev2017detecting}. \yoshiapr{The properties of the dataset are summarized in Table ~\ref{tab:datasets}.}

\vspace{-5mm}
\paragraph{Evaluation metric}
To evaluate detection performance,  we used the number of false positives per image (FPPI) and the log average miss rate (MR). 
These metrics were based on single-image detection; {\em i.e.}, they were calculated only on given test frames that were sampled discretely. Detection was performed on the given test frames and, for our method, tracking of　{ \yoshinew all} candidates was conducted in some of the subsequent frames. We used the toolkit provided for the Caltech Pedestrian Detection Benchmark~\cite{dollar2012pedestrian} to calculate the scores and plot the curves in Fig.~\ref{fig:rocdet}.

We also tested tracking accuracy separately from detection on the bird detection dataset. 
{\yoshinew We fed the ground-truth bounding boxes in the first frames to our network and other trackers, aiming to evaluate our joint detection and tracking network as a tracker.}
We conducted one-path evaluation (OPE), tracking by using ground truth bounding boxes given only in the first frame of the snippets without \yoshiaapr{re-initialization, re-detection, or trajectory fusion}. To remove very short trajectories to evaluate trackers, we selected ground truth trajectories longer than 90 frames (three seconds at 30 fps) from the annotation of the bird dataset.  We plotted success rates versus overlap thresholds. The curves in Fig.~\ref{fig:roctrk} show the proportion of the estimated bounding boxes whose overlaps with the ground truths were higher than the thresholds. 
\vspace{-5mm}
\paragraph{Object proposals}
We used a different strategy for each dataset to generate object proposals for pre-processing. In the bird dataset, we extracted the moving object by background subtraction~\cite{zivkovic2004improved}. The extracted regions were provided by the authors with the dataset; therefore, we could compare the networks fairly, regardless of the hyperparameters or the detailed tuning of the background subtraction. In the UAV dataset, we used the HOG3D-based sliding window detector provided by the authors of ~\cite{rozantsev2017detecting}.

\vspace{-5mm}
\paragraph{Compared methods}
{\it RCN~(Alex)} and {\it RCN~(VGG)} are two implementations of the
proposed method using the convolutional
layers from AlexNet~\cite{krizhevsky2012imagenet} and VGG16Net~\cite{Simonyan15}.
{\it HOG tracker+AlexNet} and {\it HOG tracker+LRCN} are baselines for the bird dataset provided by ~\cite{trinh2016}.  The former is a combination of the HOG-based~\cite{dalal2005histograms} discriminative \yoshiaapr{scale-space} tracker (DSST~\cite{danelljan2014accurate,danelljan2017discriminative}) and convnets that classify the tracked candidates into positives and negatives. The latter is a combination of DSST and the CNN-LSTM tandem model~\cite{donahue2015long}.
They used five frames following the test frames, for fair comparison, and our method used the same number of frames in the detection evaluation.

For evaluating the tracking performance, we included other combinations of the DSST and hand-crafted features for further analysis. {\it HOG+DSST} is the original version in ~\cite{danelljan2014accurate}. {\it ACF+DSST} replaces the classical HOG with more discriminative aggregated channel features~\cite{dollar2014fast}. The ACF is 
similar to HOG,
but is more powerful because of the additional gradient magnitude and LUV channels for orientation histograms. {\it Pixel+DSST} is a simplified version that uses RGB values of raw pixels instead of gradient-based features. 
\yoshiapr{ We also included ImageNet-pretrained convolutional trackers, namely, correlation-based SiamFC~\cite{bertinetto2016fully} and regression-based GOTURN~\cite{held2016learning}. They are based on the convolutional architecture \arm{of} AlexNet.} 

\begin{figure*}[t]
  \begin{center}
    \includegraphics[width=493pt]{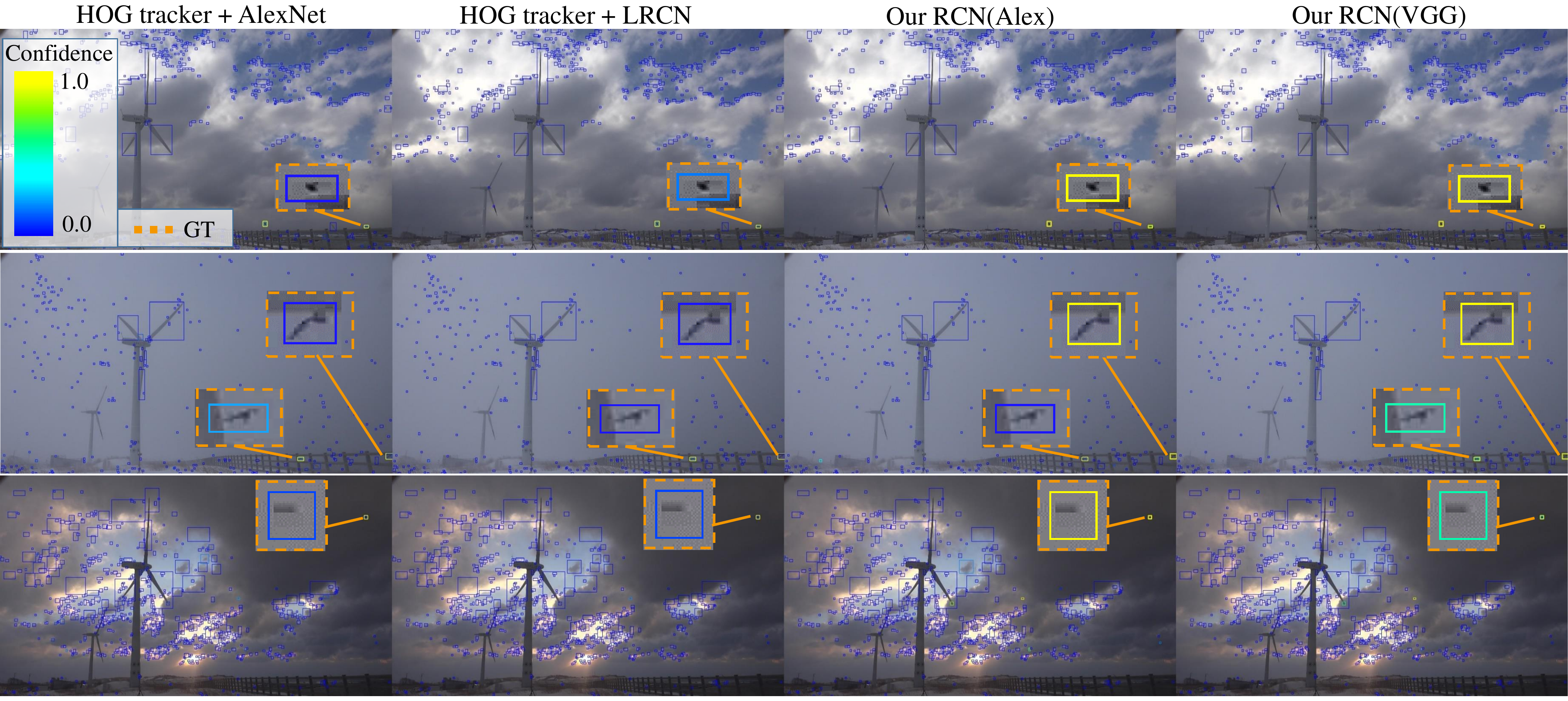}
  \end{center}
  \vspace{-7mm}
  \caption{{\yoshinew Example frames of results of detection on the bird dataset~\cite{trinh2016}. \yoshi{The dotted yellow boxes show ground truths, enlarged to avoid overlapping and keep them visible. The confidence scores of vague birds are increased and that of non-bird regions are decreased by our RCN detector. The contrast was modified for visibility in the zoomed-up samples.}}}
  \vspace{-5mm}
  \label{fig:bird_result_examples}
\end{figure*}
\vspace{-6mm}
\paragraph{Results}
The results of detection on the bird dataset are shown in Fig.~\ref{fig:rocdet}. The curves are for four subsets of the test set, which consists of birds of different sizes, namely {\it reasonable} (over 40 pixels square),  {\it small} (smaller than 40 pixels square),  {\it mid-sized} (40--60 pixels square), and {\it large} (over 60 pixels square).

On all subsets, the proposed method, {\it RCN~(VGG)} showed the smallest average miss rate (MR) of the tested detectors. 
The improvements were -10.3 percentage points on {\it Reasonable}, -2.3 percentage points on {\it Small}, 
-14.4 on {\it Mid-sized}, and -2.0 percentage points on  {\it Large} subset, in comparison with the previous best published method {\it HOG tracker+LRCN}.

A comparison of {\it HOG tracker+LRCN} and 
proposed {\it RCN~(Alex)} is also important,
because these share the same convolutional architecture.
Our {\it RCN~(Alex)} performed better on all of the subset except {\it Small}, without deepening the network.
The margins are -3.5 percentage points on {\it Reasonable}, -4.7 percentage points on  {\it Mid-sized} subset, and -0.1 percentage points on {\it Large} subset.
Examples of the test frames and results are shown in Fig.~\ref{fig:bird_result_examples} (more examples are in the supplementary material).

A comparison of {\it RCN~(Alex)} and {\it RCN~(VGG)} provides an interesting insight. {\it RCN~(Alex)} is more robust against smaller FPPI values in spite of the lower average performance than that of {\it RCN~(VGG)}. {\it RCN~(Alex)} showed a smaller MR than {\it RCN~(VGG)} when the FPPI was lower than $10^{-2}$. A possible reason is that
a deeper network is less generalizable because of many parameters; thus, it may miss-classify new negatives more often in the test set than in the shallower one.

The results of tracking on the bird dataset are shown in Fig. \ref{fig:roctrk}. We found that gradient-based features were inefficient on this dataset.  HOG-based DSST missed the target even in 30-frame short tracking (but this was already longer than in ~\cite{trinh2016} for detection). We assume that this was because of the way the HOG normalizes the gradients, which might render it over-sensitive to low-contrast but complex background patterns, like clouds. We found that replacing HOG with ACF and utilizing gradient magnitudes and LUV values benefited the DSST on the bird dataset. However, the simpler pixel-DSST outperformed the ACF-DSST by a large margin. 

The trajectories provided by our network were more robust than all of DSST variations tested. This shows that representations learned through detection tasks also work better in tracking than hand-crafted gradient features do. It also worth noting that our trajectories were less accurate than those obtained through the feature-based DSSTs when they did not miss the target.  When bounding-box overlaps larger than 0.6 were needed, the success rates were smaller than those of the DSSTs on both 30- and 60-frame tracking. This is because our network used the correlation involving pooled representation, the resolution of which was 32 times smaller than that of the original images. 
\yoshiapr{In addition, Our RCN (Alex) outperformed two existing convnet-based trackers (GOTURN and SiamFC). 
Examples of the} tracking results are presented in the supplementary material.

\begin{figure}[t]
  \begin{center}
    \vspace{-0mm}
    \hspace{-8mm}
    \includegraphics[width=260pt]{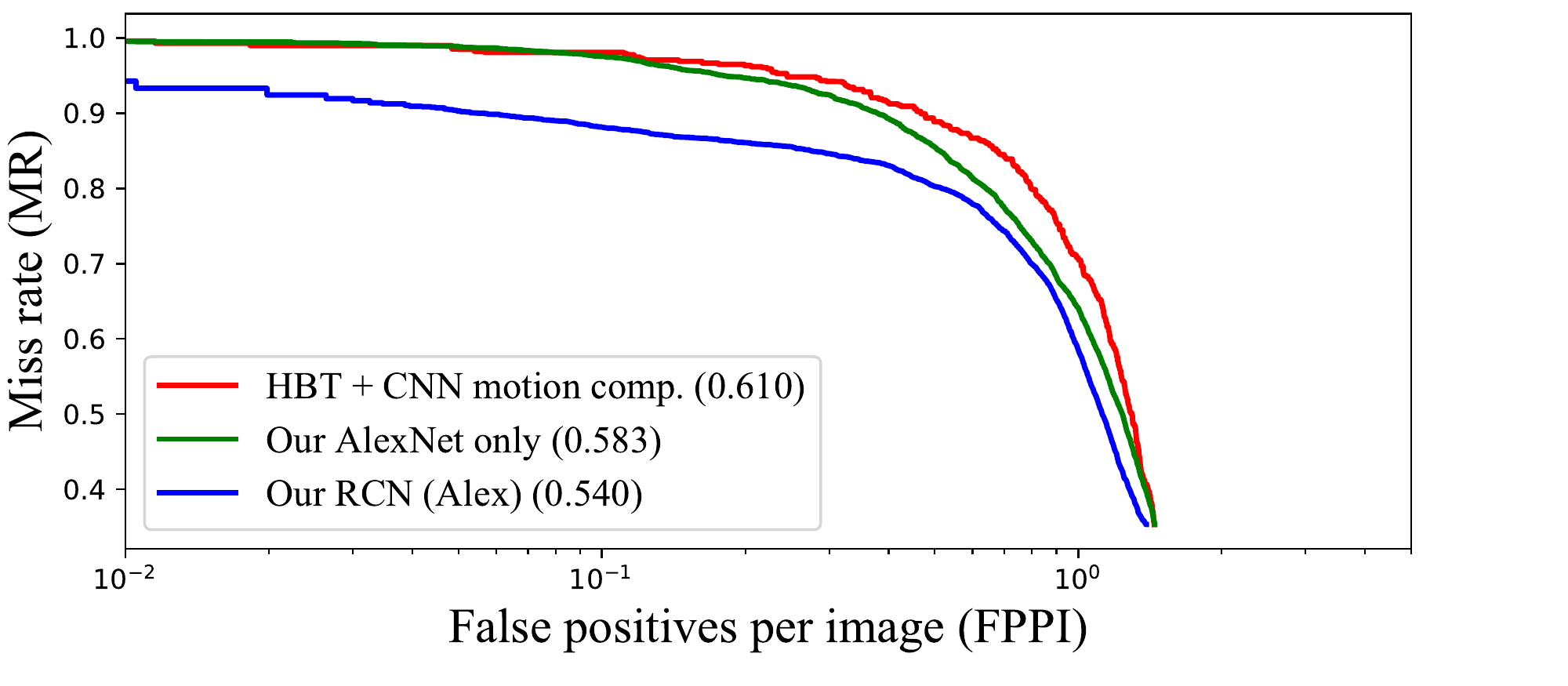}
  \end{center}
  \vspace{-8mm}
  \caption{Detection results on the UAV dataset~\cite{rozantsev2017detecting}. RCN performed the best.}
  \vspace{-6mm}
  \label{fig:roc_uav}
\end{figure}
\begin{figure}[t]
  \begin{center}
    \includegraphics[width=238pt]{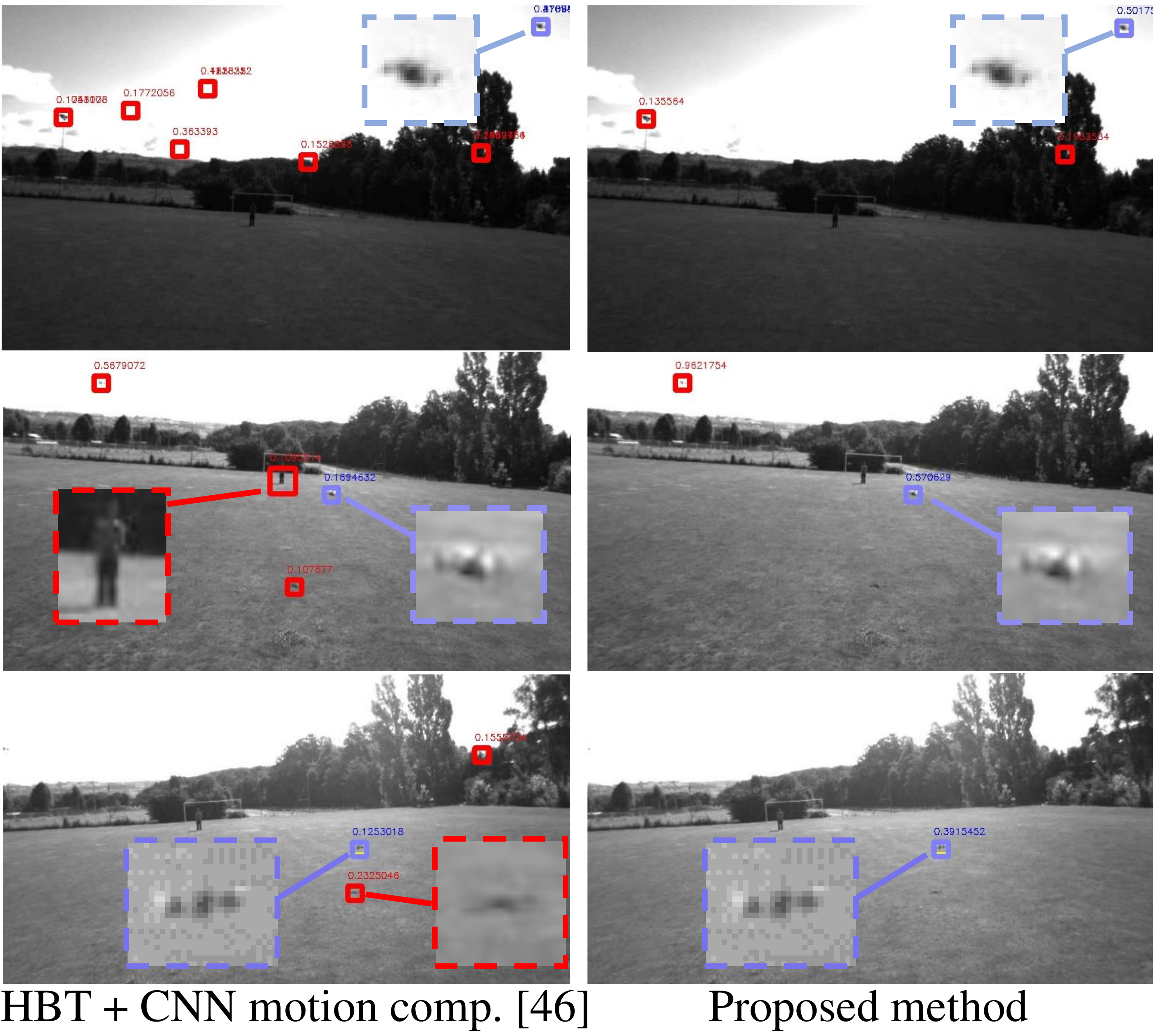}
  \end{center}
  \vspace{-6mm}
  \caption{Sample frames of detection results on the UAV dataset~\cite{rozantsev2017detecting}. The blue boxes
  show correct detections and the red ones show misdetections. \yoshi{Our method caused fewer misdetections when the detectors thresholds were set to give roughly the same MR.}}
  \vspace{-5mm}
  \label{fig:drone_samples}
\end{figure}
The resulting ROC curves of drone detection are shown in Fig.~\ref{fig:roc_uav}. We report the results of a shallower AlexNet-based version of our RCN, because of the size of the training data. We also show the curve yielded by AlexNet after single-frame pre-training without LSTM or tracking, which we refer to as {\it Our AlexNet only}. This simple implementation slightly outperformed the baseline in ~\cite{rozantsev2017detecting} without auxiliary multi-frame information by tracking or motion compensation. Our network was different in that it was deeper and larger, and had been pre-trained in ImageNet.  It is interesting that pre-training in the ImageNet classification is useful even in this domain of small, grayscale UAV detection. The ConvLSTM and joint tracking consistently improved in detection performance (-4.3 percentage points). However, the performance gain was smaller than that on the bird dataset. The reason seemed to be that the amount motion information in the UAV dataset was limited because the objects were rigid, in contrast to the articulated deformation in birds. Examples of the results are shown in Fig.~\ref{fig:drone_samples}.

\begin{table}[tb]
	\begin{center}
     \caption{Performance differences as a result of varying models and parameters. MR represents the log-average miss rate in the {\it reasonable} subset of the bird dataset, and diff. represents its difference from the baseline. $k$ denotes the kernel size of the ConvLSTM.}
     \label{tab:ablation}
	\vspace{-3mm}
    \small
  \begin{tabular}{|cl|c||c|c|} \hline
         & & Network config. & MR & diff. \\ \hline \hline
       &  \hspace{-6mm} RCN (Alex)& & & \\
       		  & k = 3 & A + B + C + D & {\bf 0.336}&   0  \\ 
    		  & k = 1 & A + B + C + D & 0.346  &   + 0.010  \\ 
              & k = 5 & A + B + C + D & 0.347  &   + 0.011 \\ \hline
              & \hspace{-6mm} RCN (VGG)& &  & \\
  	          & k = 3 & A + B + C + D & {\bf 0.268} &   0 \\ 
              & ConvGRU k = 3 & A + B + C + D & 0.271  &   + 0.003  \\ 
              & w/o tracking& A + B + D & 0.321  &   + 0.053  \\ 
              & w/o ConvLSTM& A + C + D & 0.344  &   + 0.076  \\
    		  & Single frame& A + D & 0.332  &   + 0.064  \\ \hline
              
  \end{tabular}
  \vspace{-3mm}
  \end{center}
  \vspace{-5mm}
\end{table}

\vspace{1mm}
\noindent {\bf Hyperparameters and ablation} \hspace{1mm} 
Here, we report the fluctuation on performance for different settings of the \ari{networks} and hyperparameters. We investigated the following factors: 1) kernel size in ConvLSTM, 2) ConvGRU vs. ConvLSTM, 3) w/o tracking, 4) w/o ConvLSTM, and {\yoshinew 5) single frame detection}. The kernel size controlled 
\yoshifinal{the receptive field of a memory cell.}
Second, we see the effect of simplifying ConvLSTM to ConvGRU. Third, we removed the joint tracker to see how useful multi-frame information was without stabilization. {\yoshinew Fourth, we removed the recurrent part and averaged the confidence scores through time, to see the importance of the recurrent part. Finally, we used the \yoshifinal{network} as a single-frame detector.} The results are summarized in Table \ref{tab:ablation}. \yoshinew{Here \it{Network config.} means which modules in Fig. \ref{fig:net} are active.} All of the results were in the {\it reasonable} subset of the bird dataset. The best kernel size was $k = 3$ in RCN (Alex). Larger and smaller kernels adversely affected performance \yoshinew{slightly but not critically (+0.011 and + 0.010 MR).} The performance of the ConvGRU was slightly worse than that of ConvLSTM (+0.003 MR), possibly because the input was pre-processed by convolutional layers and the burden on the recurrent part was smaller. {\yoshinew Lack of stabilization, recurrent parts, or multi-frame cues led to critical degradations in performance (+0.053, +0.076 MR and +0.064 MR), which in turn demonstrates effectiveness of the proposed network design.}
\vspace{-1mm}
\section{Conclusion} \vspace{-1mm}
We introduced the {\it Recurrent Correlation Network}, a novel joint detection and tracking framework \yoshifinal{that exploit motion information of small flying objects}.
In experiments, we tackled two recently developed datasets consisting of images of small flying objects, where the use of multi-frame information is inevitable due to poor per-frame visual information. The results showed that in such situations, multi-frame information exploited by the ConvLSTM and tracking-based motion compensation yields better detection performance. In future work, we will try to extend the framework to multi-class small object detection in videos.

\section*{Acknowledgement}
This work is in part entrusted by the Ministry of the Environment, JAPAN
(MOEJ), the project of which is to examine effective measures for preventing
birds, especially sea eagles, from colliding with wind turbines. 
This work is also supported by JSPS KAKENHI Grant Number JP16K16083, and Grant-in-Aid for JSPS Fellows JP16J04552.
The authors would like to thank Dr. Ari Hautasaari for his helpful advise to improve the manuscript's English.

{\small
\bibliographystyle{ieee}
\bibliography{mybib}
}

\end{document}